\documentclass{article}[a4]
\usepackage{authblk}

\usepackage{graphicx}
\usepackage{longtable}
\usepackage{multirow}
\usepackage[T1,hyphens]{url}
\usepackage[colorlinks,urlcolor=blue]{hyperref}

\title{Evaluation of ChatGPT feedback on {ELL} writers' coherence and cohesion}
\author[1]{Su-Youn Yoon}
\author[2]{Eva Miszoglad}
\author[3]{Lisa R. Pierce}
\affil[1]{  EduLab, Inc. \\
Tokyo, Japan \\
  \texttt{su-youn.yoon@edulab-inc.com} \\}
\affil[2]{  University of Illinois at Urbana-Champaign \\
Illinois, USA \\
  \texttt{miszogl2@illinois.edu} \\}
\affil[3]{Independent Researcher \\
Virginia, USA \\
  \texttt{lrpierce@gmail.com} \\}
\date{}

\begin{document}
\maketitle

\begin{abstract}
Since its launch in November 2022, ChatGPT has had a transformative effect on education where students are using it to help with homework assignments and teachers are actively employing it in their teaching practices. This includes using ChatGPT as a tool for writing teachers to grade and generate feedback on students' essays. In this study, we evaluated the quality of the  feedback generated by ChatGPT regarding the coherence and cohesion of the essays written by English Language Learners (ELLs) students. We selected 50 argumentative essays and generated feedback on coherence and cohesion using the ELLIPSE rubric.

During the feedback evaluation, we used a two-step approach: first, each sentence in the feedback was classified into subtypes based on its function (e.g., positive reinforcement, problem statement). Next, we evaluated its accuracy and usability according to these types. Both the analysis of feedback types and the evaluation of accuracy and usability revealed that most feedback sentences were highly abstract and generic, failing to provide concrete suggestions for improvement. The accuracy in detecting major problems, such as repetitive ideas and the inaccurate use of cohesive devices, depended on superficial linguistic features and was often incorrect.

In conclusion, ChatGPT, without specific training for the feedback generation task, does not offer effective feedback on ELL students' coherence and cohesion. 
\end{abstract}

\section{Introduction}
The advancement in natural language processing (NLP) and artificial intelligence (AI) has allowed for the development of various assistive tools for writers. The Criterion Online Writing Evaluation Service, originally designed as an automated essay scoring engine and later expanded to include writing assistance tools, and Grammarly’s grammar correction systems, stand as two notable examples. These tools have become indispensable for both native and non-native writers.

More recently, the emergence of large scale generative language models has further transformed NLP and AI. \cite{brown2020, chung2022, radford2019} demonstrated that these powerful models can learn new tasks through using only a few annotated examples, and they can achieve human-like performance on certain tasks such as text generation, question answering, and translation. Furthermore, ChatGPT, the fourth generation of the Generative Pre-trained Transformer model, has already been deployed in educational platforms and in online services such as customer support. 

ChatGPT’s ability to generate natural and human-like texts has received a lot of attention from the public and raised concerns among educators. According to one online survey, among the 1,000 U.S. college students polled, approximately 30\% of them used ChaptGPT to complete written assignments \cite{intelligenta}. This raised significant concerns regarding academic integrity, the authenticity of the students' work, and their educational progress more generally. As a result of these and related concerns, several U.S. school districts and universities have prohibited the use of AI tools in written assignments.

In a recent survey investigating teachers' use of ChatGPT \cite{intelligentb}, it was found that 97\% of the 1000 participants, which included high school teachers and university professors, responded that they have integrated ChatGPT into their teaching. They employed it for tasks such as creating lesson plans, performing grading and providing feedback, as well as composing letters of recommendation.

The application of ChatGPT for grading students' essays has begun to receive more interest. YouTube is now populated with a plethora of videos showing educators how to assess students’ essays and provide relatively ‘pain free’ feedback using ChatGPT\footnote{Example videos are \url{https://www.youtube.com/watch?v=JdrWiYmPGgw}, \url{https://www.youtube.com/watch?v=B7u0nSDKLnA}, \url{https://www.youtube.com/watch?v=RCkmo12UvPg}}. 

These video tutorials go beyond correcting grammar and suggesting vocabulary; the demonstrators are using ChatGPT to generate feedback for Higher Order Concerns (HOCs) around organization, coherence/cohesion, clarity, and content. However, unlike holistic essay grading and grammatical error corrections, which have undergone extensive evaluation studies by educational NLP researchers \cite{yancey2023, coyne2023, fang2023, wu2023}, the reliability and validity of the feedback generated by ChatGPT have not yet received systematic evaluation.

In order to address this gap, we aim to answer the following research questions:
\begin{itemize}
\item Research question 1: What approach can be used to evaluate the accuracy and usability of the feedback around coherence/cohesion?
\item Research question 2: Can ChatGPT, without the prompt and model tuning for the specific task, provide feedback around coherence/cohesion that is accurate and useful for ELL students?
\end{itemize}

The performance of ChatGPT model is known to be sensitive to the quality of the prompt\cite{reynolds2021, zhou2022}, which is a natural language instruction provided to ChatGPT for generating a response. In this study, we generate ChatGPT feedback without adapting the prompt or fine-tuning the model, because our goal is to assess the quality of the feedback that is accessible to teachers and educators who do not have computational expertise. 

\section{Literature Review}
Numerous prior studies have explored the effects of automated writing tools on formative assessment. \cite{chapelle2015,chen2008, lavolette2015} incorporated automated writing evaluation (AWE) tools, such as Educational Testing Service’s Criterion Online Writing Evaluation Service, Vantage Learning’s MY Access!, or Pearson’s WriteToLearn, in classrooms to investigate whether these tools can improve students’ writing skills.

These AWE tools typically provide holistic and analytic proficiency scores which are generated through automated scoring engines trained on a large corpus of human-scored essays and machine learning algorithms. Additionally, they offer corrective feedback for Lower Order Concerns (LOCs) such as grammatical errors, vocabulary usage, and conventions. 

While studies looking at the automatic feedback on points of LOC are common \cite{chen2008,  lavolette2015, ranalli2017}, this is counterbalanced by less information on machine generated feedback related to content and coherence. This scarcity results from the fact that automated systems offering feedback on HOCs are rare, and even when they do exist, they typically offer only a limited set of feedback functions, thus providing limited support.

For example, Vantage Learning’s MY Access! provided feedback about the organization of the students’ essays, but it relied on generic templates derived from the rubrics and lacked any personalized feedback. ETS’ Criterion Online Writing Evaluation Service and Intelligent Academic Discourse Evaluator (IADE) developed by IOWA State University provided feedback regarding the essay's discourse structure, specifically its alignment with the conventional American argumentative essay writing style known as the 5-paragraph model. This feature helped students identify problems with organization, such as a missing thesis statement or lack of a conclusion but concerns arose around its potential to encourage students to use formulaic discourse structures.

Due to these issues, many teachers’ and students’ reactions to automated feedback on the content and coherence/cohesion have been negative \cite{chen2008, dikli2010}. ESL students and instructors who took part in AWE-integrated classrooms in both studies expressed their dissatisfaction regarding the feedback due to its low accuracy, redundancy, limited coverage, and low usability. Consequently, \cite{lavolette2015} and \cite{ranalli2017} recommended using AWE tools within a limited scope, complemented by human feedback. In other words, AWE tools can be employed to provide corrective feedback for mechanical errors such as grammar and vocabulary, but teachers would still need to provide feedback for the overall organization and content. Consequently, there have been few studies regarding how to assess the feedback quality concerning content and coherence/cohesion.

Although the evaluation of content and coherence/cohesion feedback is limited, a general framework/metric was developed for assessing feedback quality. For the evaluation of the AWE systems, accuracy and usefulness of the feedback are two core metrics. These metrics have been widely used in the AWE evaluation studies\cite{dikli2010, chapelle2015, ranalli2017}. In this study, we will also use accuracy and usefulness as our main metrics.

\section{Research Question 1: What approach can be used to evaluate the accuracy and usability of the feedback around coherence/cohesion?}
\label{RQ1}
\subsection{Accuracy}
In evaluating system accuracy, precision and recall are two metrics commonly used in the NLP field. These metrics have been used to evaluate corrective feedback for grammar and vocabulary errors. Precision quantifies the ratio of true errors to the total number of errors corrected by the system. In other words, it measures how many of the corrected errors are actually true errors. On the other hand, recall measures the proportion of errors that were successfully corrected by the system out of the total number of errors present. It provides an understanding of how well the system captures and addresses all the errors. 
However, how to evaluate the accuracy of coherence/cohesion feedback has received little attention. Additionally, depending on the type of feedback, the accuracy may be difficult to judge. For instance, the template-based generic problem statements are too general to evaluate its accuracy. To address this issue, we propose a two-step approach: first perform a linguistic analysis of the feedback and evaluate the quality based on the analytic results. During the linguistic analysis, each sentence in the feedback is classified into one of the following types below, based on its function.

\cite{ellis2009} introduced a typology of corrective feedback targeting linguistic errors, such as those related to grammar and vocabulary usage. \cite{dikli2010}  extended this typology to include organization and content errors. \cite{dikli2010} categorizes the feedback into direct and indirect feedback, and the indirect feedback is subdivided into categories such as problem statement and positive reinforcement. We extend this classification by introducing two additional types: `explicit examples' and `indirect suggestions'. The final taxonomy is as follows\footnote{\cite{dikli2010} study provided the taxonomy without the definition. We provided the definition of the direct and indirect feedback from \cite{ellis2009, lyster1997}. For the remaining types, we developed the definition based on the actual examples.}:

\begin{itemize}
\item direct feedback: teachers supply the correct forms of the errors made by a student.
\item indirect feedback: teachers indicate that an error exists but do not provide the correction. For this type, the teachers can provide either the indication of the error or both indication and the exact location of the error.
\begin{itemize}
\item positive reinforcement: teachers acknowledge the correct language usage or strong points in a student's writing.

\item problem statement: teachers provide a higher-level summary about the main issues in a student’s writing.

\item explicit example: teachers identify problematic sentences (or areas) and provide examples.
\item indirect suggestion: teachers provide generic suggestions about how to solve problems in a student’s writing.
\end{itemize}
\end{itemize}

Table \ref{tab:example} provides an example of the feedback. For each sentence in the feedback, the first author annotated the feedback type based on the typology mentioned above. 

\begin{table}[htbp]
\caption{An example of feedback}
\label{tab:example}%
\begin{tabular}{|p{8cm}|l|}
\hline
Feedback Sentences    & Type      \\\hline
The essay is generally well organized with appropriate use of transitional phrases to connect ideas. & Positive reinforcement   \\\hline
However, there is some repetition and lack of variety in cohesive devices.  & Problem statement  \\\hline
For example, `Many students' is used repetitively. & Explicit example  \\\hline
The overlap of ideas is generally appropriate, & Positive reinforcement  \\\hline
but could be improved for better coherence. & Indirect suggestion \\ \hline
\end{tabular}
\end{table}

The example shows an instance of positive reinforcement, the problem statement, an explicit example, and an indirect suggestion. In this study, we distinguish the indirect suggestion from the direct feedback because the indirect suggestion tends to be overly general and lacks actionable steps, although it provides guidance about how to correct problems.

After the linguistic analysis, we evaluated the accuracy of each sentence. A comprehensive explanation about how we assessed accuracy for different feedback types will be illustrated in the Results section using  concrete examples. 

\subsection{Usefulness}
According to \cite{chapelle2015}, usefulness pertains to whether the feedback is useful for students to make decisions about revision. \cite{dikli2010, roscoe2014} evaluated the usefulness in terms of the specificity and clarity.

Specificity refers to whether the feedback provides information specific to the students’ current essay, such as explicit examples of issues and suggestions about how to correct them.

Clarity is important in ensuring that students who receive feedback can easily understand it. For example, unclear feedback includes vague expressions and undefined terminology, as demonstrated in \cite{dikli2010}, where the AWE system provides feedback to ``Use exact and specific words" without offering examples or definition about what constitutes ``exact and specific words" (p. 117). Additionally, using inappropriate or complicated language that does not match the students’ level is another instance of unclear feedback.

In assessing the usefulness of the feedback, student surveys and interviews are common methods as in \cite{ranalli2017, roscoe2014, roscoe2017}. These surveys are typically carried out after classroom studies in which students are familiarized with AWE systems and compose several essays using them. Additionally, exhaustive linguistic analysis and comparison between the system and human-generated feedback can be utilized as in \cite{dikli2010}.
In this study, we will assess the usefulness of the feedback through linguistic analysis, following a similar approach as in \cite{dikli2010} using the following procedures:
\begin{itemize}
\item Specificity: We will examine if the feedback contains information specific to the students’ current essays.
\item Clarity: We will investigate whether the feedback includes vague or unclear expressions.
\end{itemize}

\section{Research question 2: Can ChatGPT provide feedback around coherence/cohesion that is accurate and useful for ELL students?}
\label{RQ2}
In this section, we will generate coherence/cohesion feedback for a small collection of essays written by ELL students using ChatGPT. We will evaluate its quality using accuracy and usability metrics, as outlined in the preceding section.  

\subsection{Data}
In our study, we utilized a subset of data from the English Language Learner Insight, Proficiency, and Skills Evaluation (ELLIPSE) Corpus \cite{kaggle_ellips}. This publicly available corpus consists of 6,500 essays written by English Language Learners (ELLs) in the United States, spanning grades 8 to 12, and representing various levels of proficiency, ethnic backgrounds, and economic statuses. We downloaded the data from \url{https://github.com/scrosseye/ELLIPSE-Corpus/tree/main}.

The data also included holistic proficiency scores and analytic proficiency scores, which cover cohesion, syntax, vocabulary, phraseology, grammar, and conventions, as evaluated by expert raters. All the human scores in the dataset were measured on a scale of 5, with 0.5-point increments. Essays with a score of 1 were the lowest proficiency, while those with a score of 5 were the highest proficiency. 

We extracted the cohesion parameters, as the ELLIPSE rubric does not specifically include a “coherence” criteria (Presented in Table \ref{tab:rubric}), as the rubric includes both cohesion (sentence level connection) and coherence (connection of ideas). Specifically, in the ELLIPSE cohesion rubric, cohesion is assessed by looking at both cohesive devices and overall organization including logical and semantic connections. As shown in Table \ref{tab:rubric}, the organization and semantic connections were emphasized throughout all the different score levels, so we used the cohesion rubrics in our prompt to grade and generate feedback for coherence/cohesion of the essay.

\begin{table}[htbp]
\caption{ELLIPSIS rubrics for cohesion}
\label{tab:rubric}%
\begin{tabular}{|c|p{10cm}|}
\hline
{Score} & {Description} \\\hline
{1} & {No clear control of ~organization; cohesive ~devices not present
or ~unsuccessfully used; ~presentation of ~ideas unclear.} \\\hline
{2} & {Organization only ~partially developed with ~a lack of logical
~sequencing of ideas; some basic cohesive ~devices used but with
~inaccuracy or repetition. } \\\hline
{3} & {Organization generally ~controlled; cohesive ~devices used but
limited ~in type; Some ~repetitive, mechanical, ~or faulty use of
cohesive devices within ~and/or between sentences and ~paragraphs.} \\\hline
{4} & {Organization generally ~well controlled; a range ~of cohesive
devices ~used appropriately such ~as reference and ~transitional words
and ~phrases to connect ~ideas; generally appropriate overlap of
~ideas} \\\hline
{5} & {Text organization is consistently well controlled using a
~variety of effective ~linguistic features such ~as reference and
~transitional words and ~phrases to connect ideas across sentences and
~paragraphs; appropriate ~overlap of ideas.} \\\hline
\end{tabular}
\end{table}

For this study, we selected 50 essays from students in grade 12. Students answered one of the three topics about distance learning, three-year high school programs, or positive attitudes. During the essay selection process, the cohesion scores were taken into account. We decided to exclude essays with scores of .5 as the differences between essays with adjacent scores were not distinct enough to reveal clear variations in the feedback content. To achieve a balanced distribution in the scores, our initial plan was to select the same number of essays per cohesion score. However, due to the skewed distribution of the two extreme scores (1 and 5), we included all available essays for these scores, resulting in two essays with scores of 1 and three essays with a score of 5. For the remaining scores, we randomly selected 15 essays per score. Table \ref{tab:length} provides the descriptive analysis of the length of the selected essays. 

\begin{table}[htbp!]
    \caption{Length of the selected essays}
    \label{tab:length}%
   \begin{tabular}{|c|c|c|c|c|}
\hline
	& mean& standard deviation & minimum	& maximum \\\hline
number of words	&  405.0 & 	134.0	 & 205	 &  695 \\\hline
\end{tabular}
\end{table}

\subsection{ChatGPT and feedback generation method}
We used OpenAI's GPT-4 model and its text completion API to generate the feedback. The GPT-4 model received a prompt including a student’s essay as input. The GPT-4 model returned a coherence score and feedback based on the input prompt. We set the temperature parameter to zero to prevent the randomness of the output. To send the input to the GPT-4 model and receive the output, we used HTTP-based API calls instead of the playground User Interface.
The prompts consisted of a definition about the task, the rubrics for the coherence score, optional sample essays, and the input essay. The rubrics were the same rubrics presented in Table \ref{tab:rubric}.


We created two simple task definitions. While we didn't conduct a systematic prompt tuning process by optimizing the performance on the labeled output, we refined it by manually reviewing the output feedback. Initially, we collected prompts showcased in YouTube videos about the essay feedback generation. We then tested various length constraints to ensure the generated feedback conveyed comprehensive content. Additionally, we explored the different sequence of sentences and phrases to make sure that the resulting definitions could consistently produce all requested outputs (scores, general feedback, and illustrative examples from specific essays) consistently. The resulting definitions were as follows:

\begin{itemize}
\item Default: Grade the coherence/cohesion of the essay and generate feedback based on the rubrics. A score of 5 represents the highest level of proficiency. Feedback should be within 100 words.
\item Personalized: Begin by evaluating the essay’s coherence/cohesion using the rubrics, assigning a score accordingly. A score of 5 represents the highest level of proficiency. Then, provide feedback about coherence/cohesion within a limit of 100 words. Provide feedback for both major problems and strong points. When providing feedback, use specific sentences from their work as illustrative examples.
\end{itemize}

\begin{figure*}[htp]
  \centering
  \includegraphics[scale=0.85]{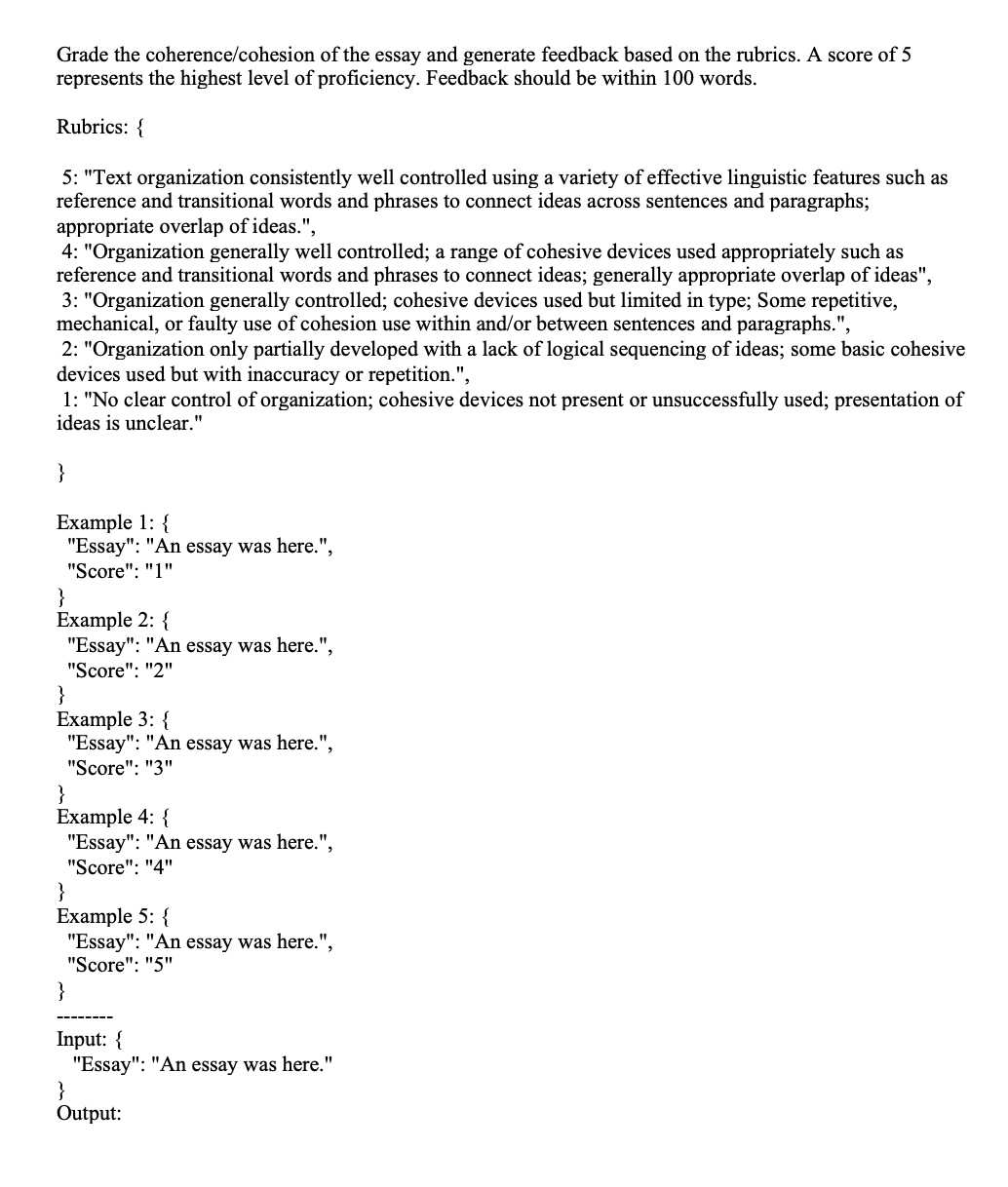}  
  \caption{Prompt with the default definition and sample essays. Because of space constraints, an actual essay was replaced with the text “An essay was here”.}
\label{fig1}
\end{figure*}

In addition, we explored the impact of the sample essays. We randomly selected one essay per score and created a Python dictionary where the score served as the key and the chosen essay as the corresponding value. For each task definition, we created two different versions of the prompt: with and without sample essays. For the prompt with the sample essays, they appeared after the rubrics. An example prompt using the default task definition with the sample essays is presented in Figure \ref{fig1}.

 \subsection{Score analysis}
Before analyzing the quality of feedback, we first evaluated the score generated by the system. The association statistics between the system and human cohesion scores were summarized in Table 
 \ref{tab:corr}.

\begin{table}[!ht]
    \centering
        \caption{exact agreement, adjacent agreement, quadratic weighted kappa, and Pearson correlation between human and system scores}
        \label{tab:corr}%
    \begin{tabular}{|p{2cm}|p{1.5cm}|p{1.5cm}|p{1.5cm}|p{1.5cm}|p{1.5cm}|}
    \hline
        Prompt type & Use of sample essay & exact agreement & adjacent agreement & quadratic weighted kappa & Pearson correlation   \\ \hline
        Default & No & 0.51 & 1.00 & 0.78 & 0.84  \\\hline
        Personalized & Yes & 0.63 & 1.00 & 0.82 & 0.86   \\ \hline
        Personalized & No & 0.59 & 1.00 & 0.81 & 0.81  \\ \hline
    \end{tabular}

\end{table}
All prompts showed strong association with the human scores; the quadratic weighted kappa ranged from $0.78$ to $0.81$ and the Pearson correlation was higher than $0.80$ for all prompts. These are substantially higher than the correlation between the number of words with the human scores of which the Pearson coefficient was 0.34. The personalized prompt (prompt extracting major problems and strength from the essay) showed better performance than the default prompt. Finally, adding sample essays dropped the performance.

In our subsequent analysis, we opted to use feedback generated from prompts without sample essays due to its superior performance in the scoring and cost-effectiveness\footnote{Users are required to pay for the utilization of the GPT-4 API, with the expenses varying based on the input length. Specifically, our costs amounted to approximately $0.003$ per essay when using a prompt without sample essays and $0.01$ per essay when using a prompt with sample essays.} because of reduced usage of tokens in the prompt. 

\subsection{Linguistic analysis of feedback}
We analyzed the length of the feedback. On average, the feedback consists of 4.1 sentences, with a range of 3 to 6 sentences. The average number of words was 62 and it ranged from 41 to 101. The prompt given to ChatGPT included the sentence that ``provide feedback within a limit of 100 word'' to manage the feedback length. The ChatGPT system adhered to this guideline, with all feedback meeting the requirement except one instance. There was no statistically significant correlation between feedback length and cohesion scores. Prior research by \cite{dikli2010} and \cite{roscoe2017} noted that the system tended to produce much lengthier feedback for students with lower proficiency, leading to their feeling overwhelmed. However, ChatGPT avoided this trend by strictly following the length limit instruction.

Next, we segmented each feedback into sentences and annotated the function of each sentence using the typology presented in the previous section. Table \ref{tab:example} presents examples of the feedback and sentence annotations. 

We analyzed the distribution of each sentence type present in the feedback to determine the predominant sentence types utilized in the feedback. The descriptive analysis of the percentage of each sentence type is presented in Table \ref{tab:sent_type_dist}. 

\begin{table}[!ht]
\caption{Percentage of sentence type in each feedback}
\label{tab:sent_type_dist}
    \begin{tabular}{|l|l|l|l|l|}
    \hline
        Sentence type & Mean & Standard deviation & minimum & maximum \\ \hline
        Direct feedback & 1 & 5 & 0 & 33 \\ \hline
        Explicit example & 31 & 16 & 0 & 67 \\ \hline
        Positive reinforcement & 22 & 20 & 0 & 67 \\ \hline
        Problem statement & 33 & 25 & 0 & 100 \\ \hline
        Suggestion & 14 & 13 & 0 & 40 \\ \hline
    \end{tabular}
\end{table}

The most frequently used sentence type was problem statements, this accounted for 33\% of the feedback, on average.  It was followed by explicit examples (31\%) and positive reinforcement (22\%). The percentage of positive reinforcement was substantially influenced by the student's proficiency levels. While the prompt was instructed to generate feedback regarding the strengths of the essay, the system struggled to provide positive reinforcement for the majority of low-scoring essays. The proportion of the positive reinforcement in the feedback provided for essays scored 1 or 2 was only 1\%. This is particularly interesting, as within our dataset, many essays with low proficiency had a clear controlling idea and arguments supporting their thesis. It is noteworthy that the ChatGPT system failed to recognize this aspect and could not generate positive reinforcement.  

The proportion of suggestion and direct feedback were low, and they were 14\% and 1\%, respectively. This pattern was in line with the current prompt design. We instructed ChatGPT to produce feedback using rubrics, where the sentences in the rubrics described general attributes for each score level. However, the rubrics did not include suggestions about how to address specific issues. We will delve deeper into this matter in the feedback-type-specific evaluation section. 

\subsection{Holistic evaluation}
In this section, we provide an evaluation of the overall feedback for an entire essay as a whole. To conduct this evaluation, two authors of this paper (hereafter, writing experts), both with extensive experience in teaching ELL students, examined the students’ essays and created their own version of the feedback. Throughout this process, we took account of the 5-paragraph model, consisting of an introduction, three body paragraphs, and a conclusion. In this model, writers are expected to introduce the topic, take a stance, and provide a preview of the arguments they are going to present in the next three body paragraphs. Each of these points need to support the controlling idea and have sufficiently explained through supporting details. The conclusion then provides an overview of all points presented in the essay and provides future implications or call to action. We paid attention to whether the students could satisfy these critical requirements and provided the feedback on their success or weakness in fulfilling them while considering ELLIPSIS rubrics.

Table \label{tab:GPT_expert_cmp} presents the examples of the ChatGPT and writing experts feedback on ELL students’ essays. We selected one example per each score, and expert feedback was based on the consensus of two authors. The essays were arranged in ascending order according to their ELLIPSIS cohesion scores. The scores for e1, e2, e3, e4 are 1, 2, 3, and 4, respectively. Each row contains complete feedback for an individual essay\footnote{Because of space constraints, we omitted the original essays of the ELL students. However, you can find instructions on how to access these essays in Appendix A.}. 

\begin{longtable}{|l|p{1.5cm}|p{7.5cm}|}
\caption{ChatGPT and writing experts feedback on students’ essays. }
\label{tab:expert_GPT_cmp}\\
\hline
Essay id & Type	 & Feedback \\\hline
\multirow{2}{*}{e1} &	ChatGPT & The essay lacks coherence and clear organization. The ideas are not logically sequenced and the use of cohesive devices is unsuccessful. For example, `I agree because is good I dea the offer of anline or video and also thus can work home' is unclear and lacks proper transition. The presentation of ideas is unclear throughout the essay.	\\\cline{2-3} 
& Expert & This essay has several typos which make reading hard, so clarity needs work. In terms of coherence, the controlling idea of the essay is quite clear and there are arguments supporting the thesis, however the supporting details are not sufficient and need further progression.\\ \hline
\multirow{2}{*}{e2} & ChatGPT &	The essay lacks logical sequencing of ideas and uses basic cohesive devices inaccurately. The writer jumps from one idea to another without clear transitions. For example, the shift from discussing cheating to distractions at home is abrupt. The essay could benefit from more effective use of transitional words and phrases to connect ideas. \\\cline{2-3}	 
& Expert & It is hard to understand the essay's points as the first paragraph is an introduction to the topic and the second paragraph is about all the negative elements of distance learning. The third paragraph however is unrelated and somewhat contradictory to the previous paragraph. The essay simply does not contain sufficient information to support the points. I understand the central idea but the supporting details are lacking which makes the essay weak. Each paragraph should have a clear topic sentence with the support as this essay jumps from one thing to the next. \\ \hline
\multirow{2}{*}{e3} &	ChatGPT & The essay has a clear structure with an introduction, body, and conclusion. However, the use of cohesive devices is limited and sometimes repetitive. For example, `That is why communication so important that distance learning can not offer to students.' could be better connected to the previous sentence. Also, the essay could benefit from more transitional words and phrases to connect ideas across sentences and paragraphs. \\\cline{2-3} 
& Expert & Lexicogrammar is simple with noticeable errors that cause processing difficulty. There is a clear structure with 5 paragraphs with a clear controlling idea. Main points are: lack of communication access, lack of competition, no life skills/inspiration from teachers. This essay is actually great in terms of content, but the clarity of language needs work including grammar. \\\hline
\multirow{2}{*}{e4} & ChatGPT &	The essay is generally well organized with appropriate use of transitional phrases to connect ideas. However, there is some repetition and lack of variety in cohesive devices. For example, `Many students' is used repetitively. The overlap of ideas is generally appropriate, but could be improved for better coherence. \\\cline{2-3}
& Expert &	The controlling idea of the whole essay is never explicitly stated. The main points are transportation, not attending school everyday is good, being in school is good, technology advancement: school can be online--some contradiction which obstructs the flow of the paper. Grammar is OK, but the organization of points and clear logic are missing. \\\hline
\end{longtable}

As outlined above in the 5-paragraph model, cohesion needs to be evaluated based on essays having one central idea that is developed throughout the essay. All essays discussed in Table \ref{tab:expert_GPT_cmp} were coherent in that they took a stance and provided supporting ideas. But two low score essays (e1 and e2) lacked effective paragraph development and failed to fully explain and support their view using sufficient examples and explanations. Notably, ChatGPT failed to recognize the presence of a central idea in these two low-scoring essays. This failure led to distinct differences in the feedback provided by the writing experts and ChatGPT. The latter identified the primary issues as a lack of coherence and clear organization in both essays, but the writing experts focused their feedback on insufficient supporting evidence and lack of paragraph development.   

All the ChatGPT feedback in Table \ref{tab:expert_GPT_cmp} included comments on the use of the cohesive devices. For the four essays discussed in Table \ref{tab:expert_GPT_cmp}, ChatGPT commented on inaccurate usage, repetition, and a lack of diversity.  Beyond these four essays, all ChatGPT feedback generated in this study included the comments regarding the cohesive devices or transitional words. However, we found that these comments were not always accurate as some of the repetitions were related to the essay topic and were conducive to the progression of ideas. Additionally, it should be noted that the feedback about the cohesive device may be less pertinent for the essays at lower proficiency levels, as illustrated by two lower-scoring essays in Table \ref{tab:expert_GPT_cmp}. In these cases, the cohesive device was not the primary issue, and following the feedback and adding more transitions would not improve the clarity of these two essays. We will further discuss this point with the detailed analysis in the following section.

\subsection{Feedback-type-specific evaluation}
In this section, we will classify each sentence into the different type based on the typology presented in the previous section and evaluate the accuracy and usability of the feedback for each type. 

First, we will evaluate the positive reinforcements and problem statements. Table \ref{tab:PR_PS_rubric} presents the examples of the the positive reinforcements and problem statements generated by ChatGPT. For each ChatGPT sentence, we also present the semantically closest sentence from the ELLIPSIS cohesion rubrics (`Similar rubric sentence' column). If there is no similar sentence in rubrics, it is marked by ``no similar sentence''. In addition, the score that the selected rubric sentence belonged to is presented in the parenthesis. For instance, the feedback presented in the first row ``the essay lacks coherence and clear organization'' was similar to the rubric sentence ``No clear control of organization'' and this sentence described the score of 1.

\begin{longtable}{|l|p{5cm}|p{5cm}|}
    \caption{Examples of positive reinforcement and problem statement}
    \label{tab:PR_PS_rubric} \\
    \hline
        Score & Feedback sentence & Similar rubric sentence (score from the rubric) \\ \hline
        \multirow{2}{*}{1} & the essay lacks coherence and clear organization & No clear control of  organization (score 1) \\ \cline{2-3}
         & the ideas are not logically sequenced and the use of cohesive devices is unsuccessful & Organization only partially developed with a lack of logical sequencing of ideas (score 2)  \\ \hline
        \multirow{2}{*}{2} & the essay lacks coherence due to poor sentence structure and grammar & No similar sentence \\ \cline{2-3}
         & the ideas are not logically sequenced and the use of cohesive devices is inaccurate & Organization only partially developed with a lack of logical  sequencing of ideas; some basic cohesive devices used but with  inaccuracy or repetition. (score 2)  \\ \hline
        \multirow{3}{*}{3} & the essay has a clear structure with an introduction, body, and conclusion & No similar sentence \\ \cline{2-3}
        &  the coherence is affected by repetitive and mechanical use of phrases & cohesive devices used but limited in type; Some repetitive, mechanical, or faulty use of cohesion (score 3)  \\ \cline{2-3}
       &  the essay also lacks transitional words and phrases to connect ideas & cohesive devices not present or unsuccessfully used (score 1)  \\ \hline
        \multirow{2}{*}{4} & the essay is generally well organized with appropriate use of transitional words and phrases to connect ideas  & Organization generally well controlled; a range of cohesive devices used appropriately such as reference and transitional words and phrases to connect idea (score 4) \\ \cline{2-3}
        &  there is some repetition and lack of variety in cohesive devices & Some repetitive, mechanical, or faulty use of cohesion use within and/or between sentences and paragraphs (score = 3) \\ \hline
\end{longtable}

In ChatGPT feedback, the same sentences recurred repetitively; many sentences in the feedback column of Table \ref{tab:PR_PS_rubric} appeared in several different essays.  The majority of the positive reinforcements or problem statements were copies or slight modification of rubrics. In Table \ref{tab:PR_PS_rubric}, 6 out of 8 ChatGPT sentences had matching sentences in the rubrics. In addition, we calculated the text similarity between the rubric sentences and ChatGPT feedback using the entire samples used in this study.  The overlap of content words reached 79\%, signifying a significant level of content similarity. This result implies that the  additional information, generated by ChatGPT for each individual essay, conveyed by positive reinforcements and problem statements, beyond what is already present in the rubrics, is not extensive.

Furthermore, the problem statements were not always accurate or did not reflect the reason for the lack of coherence for the specific essay. There were multiple instances where rubric sentences that describe essays with different scores were included in the feedback. For instance, the ChatGPT used ``the essay also lacks transitional words and phrases to connect ideas'' for several essays with score 3. This feedback  was similar to the rubric sentence ``cohesive devices not present or unsuccessfully used'' which was the description of score of 1. 

In order to provide a clear analysis of this issue, we selected two essays and provided its whole ChatGPT feedback. The feedback for each essay was presented in Table \ref{tab:exmple2}: 

\begin{table}
\caption{Two examples of ChatGPT feedback}
\label{tab:exmple2}
\begin{tabular}{|l|p{10cm}|}
\hline
 Essay id & ChatGPT feedback \\\hline
 e5 & The essay has a clear structure with an introduction, body, and conclusion. However, the use of cohesive devices is limited and repetitive. For example, ``we can sleep more'' is repeated without variation. The essay also lacks transitional words and phrases to connect ideas. The presentation of ideas is generally controlled but could be improved with more varied and effective linguistic features. \\\hline
 e4  & The essay is generally well organized with appropriate use of transitional phrases to connect ideas. However, there is some repetition and lack of variety in cohesive devices. For example, ``Many students'' is used repetitively. The overlap of ideas is generally appropriate, but could be improved for better coherence. \\\hline
\end{tabular}
\end{table}

In the feedback for essay e5, ChatGPT had two comments about the cohesive devices. They were ``However, the use of cohesive devices is limited and repetitive'', and ``the essay also lacks transitional words and phrases to connect ideas''. Notably, the essay itself incorporated connective words like ``First", ``Then'', ``Finally'', and ``In conclusion,'' rendering this problem statement inaccurate. A more critical issue was that the problem statement failed to address the core reason behind the lack of coherence. The issue was not in the repetition of the phrase ``we can sleep more'', but rather in two paragraphs addressing the same points without a progression of ideas to advance the essay's controlling idea. The ChatGPT system failed to recognize the main issue and provided feedback on a superficial error. 

In the feedback for essay e4, the feedback included the problem statement that ``there is some repetition and lack of variety in cohesive devices''. This problem statement sentence was used imprecisely. This essay generally lacked coherence because there was a contradiction within the essay that made reading it difficult and problematic. On the contrary, it used a few cohesive devices without repetition. Therefore, the problem was in the contradiction of the ideas, not in the cohesive device usage. In this case, the student needed to take a firm stance and support the arguments throughout the essay rather than incorporating or varying cohesive devices.

The problem statements and positive reinforcements generated by ChatGPT were generic and abstract making it challenging to ascertain the precision of the feedback across many instances. In addition, the utility of these statements remained quite limited. Both specificity and clarity were notably low.

Next, we evaluated the explicit examples. On average, approximately 1.21 sentences of explicit example were included in the individual feedback for each essay. Most feedback included one explicit example, while some feedback included three explicit examples.

The explicit example provided the problematic sentences extracted from the specific essay. We evaluated whether these sentences were generated by ChatGPT as it is well-known that it can fabricate content (hereafter, hallucination error). We found that all the sentences included in the explicit example were actually part of the respective essays, and there was zero hallucination error. 

Qualitative analysis of the explicit examples in this study revealed three major problems: inaccurate grading about repetitive ideas, transitional word usage, and unclear ideas. Table \ref{tab:exp_repetition} provides instances  of explicit examples about the repetitive ideas. Each row was a part of the feedback for a distinct essay.

\begin{table}[htbp]
\caption{Explicit example about repetition of ideas}
\label{tab:exp_repetition}
\begin{tabular}{|l|p{10cm}|}
\hline
Essay id &	Explicit example \\\hline
e4 & for example ``many students'' is used repetitively \\\hline
e6 & for example ``online claases are not like regular claases you take'' is repeated without adding new information \\\hline
e7 & for example ``i wanna talk about'' is used repetitively to introduce new ideas \\\hline
e5	& for example ``we can sleep more'' is repeated without variation \\\hline
\end{tabular}
\end{table}

The ChatGPT system detected the repetition only when the exact words or phrases appeared multiple times. In certain instances, reiteration of topic-related terms (e.g., online classes) was identified as an error, although experts viewed it as innocuous and not a significant concern, often due to writers' limited vocabulary, while still attempting to show interconnectedness of body paragraph points to the thesis of their essay. Conversely, it struggled to discern repeated ideas or concepts when they were expressed using different wording. In some instances, these reiterated concepts spanned several paragraphs within an essay, yet no feedback was provided concerning this matter.

Table \ref{tab:exp_transition} provides examples of the explicit example about the use of the transitional words.

\begin{table}[htbp]
\caption{Explicit example about transitional word use }
\label{tab:exp_transition}
\begin{tabular}{|l|p{10cm}|}
\hline
Essay id& Explicit example \\\hline
e8 & the use of transitional phrases like ``one'', ``two'', and ``three'' helps to connect ideas across paragraphs  \\\hline
e9 & the use of transitional phrases like ``secondly'', ``on the other hand'', and  ``in addition'' helps to connect ideas  \\\hline
\end{tabular}
\end{table}

As shown in Table \ref{tab:exp_transition}, the ChatGPT system had a tendency to over-reward the use of simplistic cohesive devices, for example, the use of connectives such as ``first'', ``second'', and ``in conclusion''. These were highly evaluated and resulted in positive feedback even if the transitions between sentences were not smooth and there were gaps in the ideas. Conversely, the essays that connected ideas smoothly without these proto-typical connectives tended to receive negative feedback about the ``absence of transitional words''.

The usability of feedback was relatively high regarding the transitional words and repetition of ideas; the specificity was high because the feedback encompassed the problematic sentences extracted from the specific essay. In addition, understanding the problems and correcting them were relatively easy although the suggestions about how to correct them were not explained in the feedback. For instance, for the feedback ``for example  `many students' is used repetitively'', the student needed to replace the mentioned phrase with different phrases or pronouns. Nonetheless, the feedback had a low level of accuracy, which led to the quality of the corrected essays following the feedback remaining unchanged.

Finally, Table \ref{tab:exp_unclear} provides examples of the unclear ideas. 
\begin{table}[htbp]
\caption{Explicit example about unclear ideas}
\label{tab:exp_unclear}
\begin{tabular}{|l|p{10cm}|}
\hline
Essay id &	Explicit example \\\hline
e10	& for example ``when the people decide to have a good posture in they live of thing was be better'' is unclear and lacks coherence  \\\hline
e11	 &	for example ``if you image the world student stay home take online class'' is unclear and lacks connection to the following sentences  \\\hline
e12	 &	for example ``may mean idea is students is coming the different countries is heart because speak different languages y different cultures is different languages the people is coming and this countries because make you future for to pass away you life'' is difficult to understand due to lack of clear organization and coherence  \\\hline
e13	&	for example ``if you take a decision studied studied because in your future is helping the stuied is good'' is unclear and lacks connection to previous ideas  \\\hline
\end{tabular}
\end{table}

The ChatGPT system successfully found sentences including severe errors that impede understanding. When identifying these problematic sentences, the ChatGPT system primarily relied on grammar and vocabulary errors, resulting in overlooking other types of unclear sentences that exhibited proper grammar but unclear meaning or unconnected sentences. In summary, the ChatGPT system showed high precision but low recall in detecting unclear sentences. This could potentially lead to a situation where essays with extensive syntactic and grammatical errors are penalized excessively, while essays with semantic issues receive inflated scores. Furthermore, even if the accuracy of these feedback may be relatively higher than feedback about other issues, the utility of the feedback, particularly in terms of clarity, remained limited. The feedback mentioned the problematic sentences without providing explanations for the issues or offering suggestions for sentence improvement.

\section{Discussion}

During our analysis of ChatGPT feedback, we observed that it primarily assessed the surface-level use of cohesive devices; It appeared to be sensitive to the presence or absence of transitional words, but it could not distinguish inefficient and inaccurate usage from efficient and accurate usage. As an exercise, we chose 4 exemplar essays, one in each score range, made minor modifications in the transitional words in the original essays, and regenerated the score and feedback using the same prompt we used to generate the initial feedback. The modified essays were reviewed by the writing experts who confirmed that the modifications did not significantly alter the quality of the essays.

Table \ref{tab:regenerated_feedback} presents the essay id, cohesion score, changes made in the original essay, the initial feedback, and regenerated feedback for the modified essays. The original ChatGPT feedback is presented in Table \ref{tab:expert_GPT_cmp}.

\begin{longtable}{|l|p{3cm}|p{6cm}|}
    \caption{Regenerated ChatGPT feedback for the modified
    essay}
\label{tab:regenerated_feedback} \\
    \hline
Essay id & Modification made to the original essay & Regenerated feedback \\\hline
e1 & Insertion of three connectives (``first'', ``second'', ``finally'' at the beginning of  each paragraph in the body) & No change in feedback \\\hline
e2 & Insertion of four connectives (``first'', ``second'', ``finally'', ``in conclusion'' at the beginning of each paragraph in the body) & The essay lacks logical sequencing of ideas and uses basic cohesive devices inaccurately. The transition between paragraphs is abrupt and lacks coherence. For example, the transition from ``Second, if they can not go to school...'' to ``Finally, if the student have a little brother or sister...'' is disjointed. The essay could benefit from more effective use of transitional words and phrases to connect ideas across sentences and paragraphs. \\\hline
e3 &  Removal of three connectives (``the first'', ``the second'', ``the last'', ``conclusion'' from the beginning of each paragraph in the body & The essay has a clear structure and presents ideas in a logical order. However, the use of cohesive devices is limited and sometimes repetitive. For example, ``That is why communication so important that distance learning can not offer to students.'' could be better connected to the previous sentence. Also, the transition between paragraphs could be smoother. The essay would benefit from more varied transitional words and phrases. \\\hline
e4 & Removal of one connective (``in my conclusion'') from the conclusion paragraph & The essay is generally well organized with a clear progression of ideas. Cohesive devices such as reference and transitional words are used appropriately, for example, ``One impediment may be transportation'', ``As we see many students have a job after school'', ``It is a great option to have schools that offers distance learning''. However, there is room for improvement in the use of transitional phrases to connect ideas across paragraphs. \\\hline
\end{longtable}

The system scores remained unchanged despite modification. Our initial thought was that ChatGPT might increase the coherence scores as a result of adding transition words - since that was ChatGPT’s recommendation - to use more cohesive devices. The insertion or deletion of the transitional words neither inflated, nor deflated the scores, however, the feedback still focuses on the use of cohesive devices only this time suggesting more varied use of them.  

It is reassuring that the score of ChatGPT remained stable for simple insertion of the connectives. However, it is important to acknowledge that our experiment was limited and the size of the test samples was only four.  Assessing ChatGPT’s resilience against the gaming attempts requires extensive and comprehensive future studies. Nonetheless, it is noteworthy that following the feedback and simply inserting connectives would not improve both the quality of the essay and their scores could potentially lead students to frustration and distrust to the system, as following ChatGPT’s feedback for revision will still not improve scores. 

\section{Conclusions}

In this study, we investigated the quality of the ChatGPT feedback on coherence/cohesion of ELL students’ essays. To achieve this goal, we first developed metrics to assess the quality of the feedback. Our approach focused on evaluating accuracy and usability and incorporated feedback-type-specific evaluation steps. For the feedback-type-specific evaluation, we augmented  \cite{dikli2010}’s feedback typology and developed a comprehensive guideline for assessing the quality of each feedback type. 

Subsequently, we generated coherence/cohesion scores and feedback using ChatGPT for 50 essays written by ELL students and evaluated them. First, we evaluated the quality of the scores by calculating the agreement statistics with human scores. ChatGPT scores showed strong association with the human scores, with both correlation and weighted kappa for the best model surpassing 0.80. When comparing prompts with and without sample essays, the inclusion of sample essays did not lead to additional improvements. 

The comparison with the feedback generated by writing experts, followed by a detailed analysis specific to feedback types, revealed a few major issues in the accuracy of the feedback. One significant problem was ChatGPT's inability to identify the main controlling ideas in lower-scoring essays, leading to inaccurate assessments of their overall organizational quality. Moreover, the comments about cohesive devices were included in the majority of the feedback, but the usage was assessed imprecisely, as it heavily relied on the presence or absence of connectives rather than their efficacy or appropriateness. Additionally, the detection of the repeated ideas also relied on the superficial word matching, thus failing to identify substantial idea repetitions expressed in different words or phrases were undetected. As a result, considerable parts of the feedback were inaccurate, and some major issues remained unaddressed. 

Our analysis of the feedback distribution showed that the majority of sentences were problem statements and positive reinforcements, while direct feedback was rare. The problem statements and positive reinforcements were highly abstract and did not provide any concrete suggestions for improvement. Furthermore, the word similarity analysis between ChatGPT feedback and ELLIPSIS Rubrics uncovered that ChatGPT used almost exclusively the rubric’s language and offered not much new information. In summary, the feedback did not provide students with sufficient explanation of the assessment they receive and how they can correct and improve their writing, and the usability of the feedback was low. 


We concluded that ChatGPT without prompt or model tuning on the specific task does not provide effective feedback on ELL students’ coherence/cohesion. Using ChatGPT alone would not provide students with the feedback necessary to make adjustments and meaningful revisions, as ChatGPT feedback is too general and does not provide actionable or global implications about their writing. We do not recommend the use of ChatGPT to instructors to assist with evaluating ELL writers' coherence/cohesion unless instructors adapt the ChatGPT system for the specific task. While ChatGPT can effectively correct lower order concerns, we recommend that instructors address higher order concerns in ELL writing. 

However, our conclusions are restricted to the ChatGPT without training for the coherence/cohesion feedback task. ChatGPT's performance is significantly influenced by the prompts, and prompts optimized for coherence/cohesion feedback generation might achieve higher accuracy and enhanced usability. According to our analysis, it would be beneficial to initially create models capable of accurately identifying fundamental issues related to coherence and cohesion. These issues include recognizing controlling ideas, identifying repetitive concepts, and distinguishing between inappropriate and correct usage of transitional words. Improvements in detecting these fundamental issues, along with generating clear and actionable suggestions about these problems, may result in the generation of meaningful and valuable feedback for coherence/cohesion.

Finally, it is important to note that these results were based on a limited number of test essays and a single set of samples. In future research, we plan to use a larger data set with more varied sample essays and to expand our inquiries around feedback quality to include scoring accuracy.  

\section{acknowledgement}
We extend our gratitude to the Learning Agency team, with special thanks to Ulrich Bose, Alex Franklin, and Perpetual Baffour, for generously sharing the ELLIPSIS dataset and rubrics, as well as for their prompt and helpful responses to our inquiries.

\bibliographystyle{plain}
\bibliography{organization}

\appendix 
\section{Accessing the students' essays}

You can access all the students' essays at the following URL: \url{https://raw.githubusercontent.com/scrosseye/ELLIPSE-Corpus/main/ELLIPSE_Final_github.csv}.

To find the particular essays referenced in this paper, you'll need to cross-reference our essay id with the ELLIPSIS id in Table \ref{tab:map}. Once you have the ELLIPSIS id, you can use it to locate the corresponding student's essay in the provided CSV file.

\begin{table}[htbp]
\caption{Table between essay id in this paper and the original ELLIPSIS id}
\label{tab:map}
\begin{tabular}{|l|l|}
\hline
Essay id &	ELLIPSIS id \\\hline
\hline 
e1 & C716B4A085B3 \\\hline 
e2 & 1BA83A25F117 \\\hline
e3 & 70041E13A0CC \\\hline
e4 & 02B36EBD5C66 \\\hline
e5 & 292A64D51CE3	\\\hline
e6 & 1E563F170C61 \\\hline
e7 & 21460A7F2719 \\\hline
e8	& 214EC21D1403 \\\hline
e9	& 2652194CF346 \\\hline
e10	& 01350DF42AED \\\hline
e11	& 04E1A9EE5BE1 \\\hline
e12	& 0C8D699DEE90 \\\hline
e13 & 0F84CCFB4982 \\\hline
\end{tabular}
\end{table}

\end{document}